\documentclass[letterpaper, 10 pt, conference]{ieeeconf} 

\IEEEoverridecommandlockouts    

\makeatletter 
\newcommand{\removelatexerror}{\let\@latex@error\@gobble}
\makeatother

\usepackage[ruled, vlined]{algorithm2e}
\usepackage{amsfonts}
\usepackage{amsmath}
\usepackage{amssymb}
\usepackage{amsthm}
\usepackage{array}
\usepackage{epsfig}
\usepackage{epstopdf}
\usepackage{graphicx}
\usepackage{placeins}
\usepackage{subfigure}
\usepackage{wrapfig}
\usepackage{url}

\graphicspath{{figures/}}

\title{\LARGE \bf Pick and Place Without Geometric Object Models}
\author{Marcus Gualtieri, Andreas ten Pas, and Robert Platt
\thanks{College of Computer and Information Science, Northeastern University, Boston, Massachusetts, USA. This work has been supported in part by NSF through IIS-1427081 and ONR through N000141410047.}
}

\begin{document}

\maketitle
\thispagestyle{empty}
\pagestyle{empty}

\begin{abstract}

We propose a novel formulation of robotic pick and place as a deep reinforcement learning (RL) problem. Whereas most deep RL approaches to robotic manipulation frame the problem in terms of low level states and actions, we propose a more abstract formulation. In this formulation, actions are target reach poses for the hand and states are a history of such reaches. We show this approach can solve a challenging class of pick-place and regrasping problems where the exact geometry of the objects to be handled is unknown. The only information our method requires is: 1) the sensor perception available to the robot at test time; 2) prior knowledge of the general class of objects for which the system was trained. We evaluate our method using objects belonging to two different categories, mugs and bottles, both in simulation and on real hardware. Results show a major improvement relative to a shape primitives baseline.

\end{abstract}

\section{Introduction}
\label{sec:introduction}

Traditional approaches to pick-place and regrasping require precise estimates of the shape and pose of all relevant objects \cite{LozanoPerez1986,Mason1985}. For example, consider the task of placing a mug on a saucer. To solve this problem using traditional techniques, it is necessary to plan a path in the combined space of the mug pose, the saucer pose, and the manipulator configuration. This requires the pose and shape of the mug to be fully observable. Unfortunately, even when the exact shape of the mug is known in advance, it can be hard to estimate the mug's pose precisely and track it during task execution. The problem is more difficult in the more realistic scenario where the exact shape of the mug is unknown.

Approaches based on deep RL are an alternative to the model based approach described above~\cite{Mnih2015}. Recent work has shown that deep RL has the potential to alleviate some of the perceptual challenges in manipulation. For example, Levine \textit{et al.} showed deep learning in conjunction with policy gradient RL can learn a control policy expressed directly in terms of sensed RGB images~\cite{Levine2016A}. Not only does this eliminate the need to develop a separate perceptual process for estimating state, but it also simplifies the perceptual problem by enabling the system to focus on only the perceptual information relevant to the specific manipulation task to be solved. This, along with encoding actions using low level robot commands (such as motor torque or Cartesian motion commands~\cite{Levine2016B,Levine2016A}), means the approach is quite flexible: a variety of different manipulation behaviors can be learned by varying only the reward function.

Unfortunately, deep RL approaches to robotics have an important weakness. While the convolutional layers of a deep network facilitate generalization over horizontal and vertical position in an image, they do not facilitate generalization over depth or in/out of plane orientation, i.e., the full 6-DOF pose space in which robots operate. This is a significant problem for robotics because  deep RL methods must learn policies for many different relative poses between the robot and the objects in the world. Not only is this inefficient, but it detracts from the ability of the deep network to learn other things like policies that generalize well to novel object geometries.

\begin{figure}[t]
    \centering
    \includegraphics[height=1.3in,trim={6in 4in 2in 0in},clip]{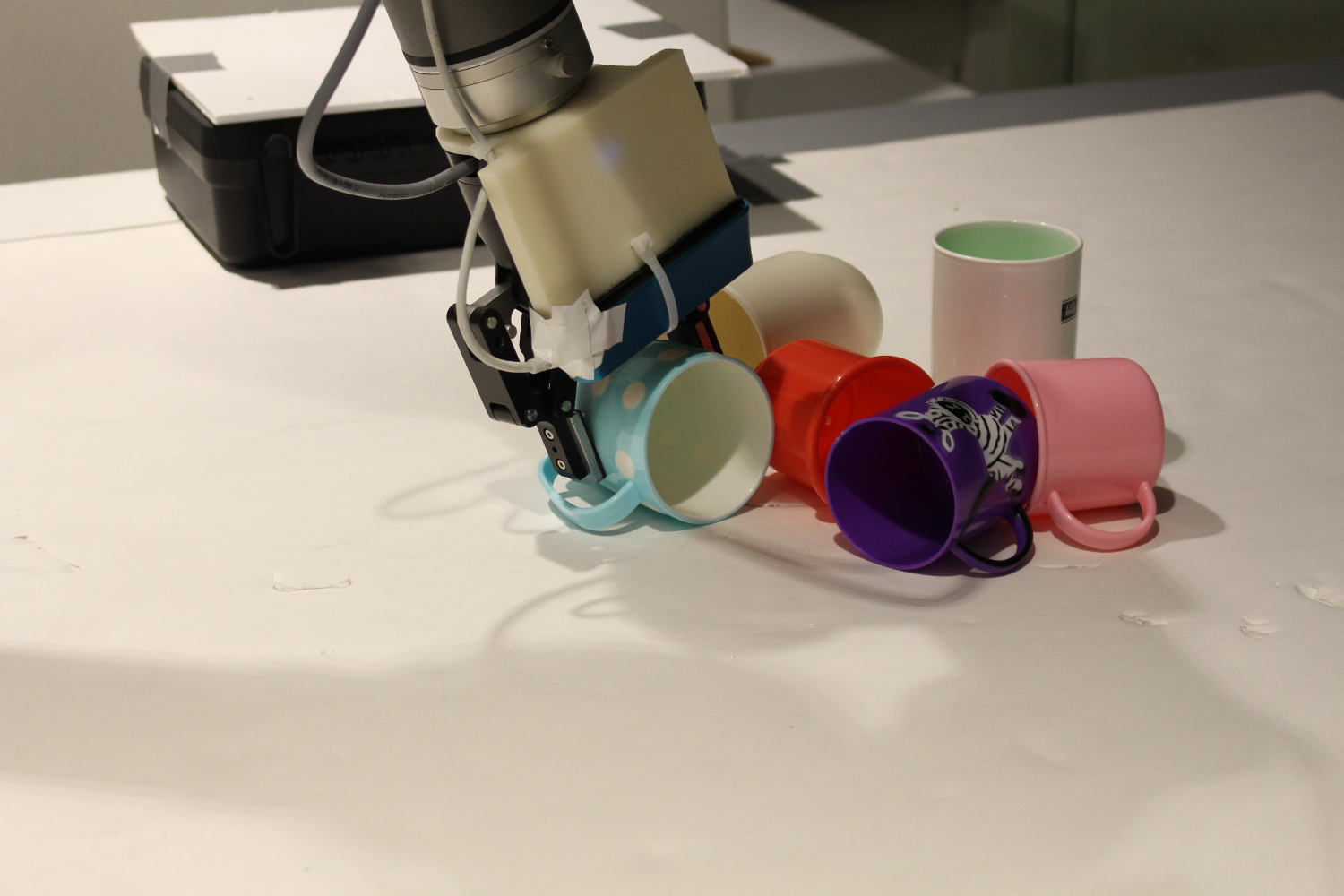}
    \includegraphics[height=1.3in,trim={2.5in 5in 7in 0in},clip]{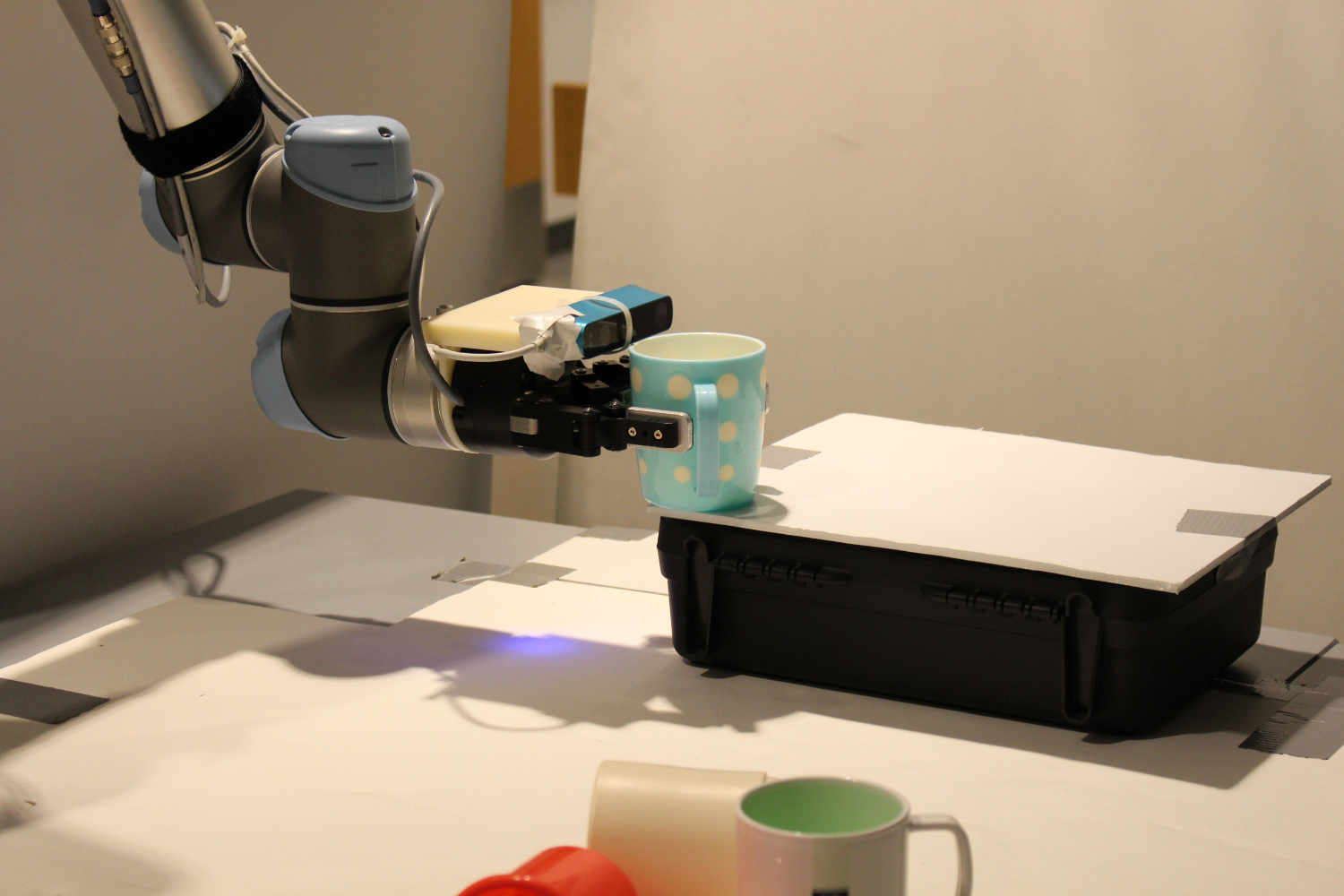}
    \caption{Pick-place problem considered in this paper. The robot must grasp and place an object in a desired pose without prior knowledge of its shape.}
    \label{fig:illustratePickPlace}
    \vspace{-0.2in}
\end{figure}

We propose a new method of structuring robotic pick-place and regrasping tasks as a deep RL problem, i.e., as a Markov decision process (MDP). Our key idea is to formulate the problem using reach actions where the set of target poses that can be reached using these actions is sampled on each time step. Each reach action is represented by a descriptor that encodes the volumetric appearance of the scene in the local vicinity of the sampled reach target. In order to formulate the MDP, we note our problem is actually a partially observable MDP (POMDP) where object shape and pose are hidden state and the images or point clouds produced by the sensors are the observations. In order to solve this problem as an MDP, we encode belief state as a short history of recently taken reach actions expressed using the volumetric descriptors used to encode the reach action. 



As a result of these innovations, our method is able to learn policies that work for novel objects. For example, we show that our system can learn to grasp novel mugs (for which prior geometric models are not available) from a pile of clutter and place them upright on a shelf as in Figure~\ref{fig:illustratePickPlace}. The same system can be trained to perform a similar task for other classes of objects, such as bottles, simply by retraining. Our system can also learn policies for performing complex regrasping operations in order to achieve a desired object pose. As far as we know, this is the first system described in the literature that has been demonstrated to accomplish the above without constructing or matching against geometric models of the specific objects involved.

\section{Related Work}
\label{sec:relatedWork}

One early approach to manipulation of unknown objects is based on shape primitives. Miller \textit{et al.} explored this in the context of grasp synthesis~\cite{Miller2003}. Others have extended these ideas to segmentation and manipulation problems~\cite{Rusu2009,Morwald2010,Harada2013}. These methods have difficulty when the objects are not approximately cylindrical or cuboid and when the objects cannot be easily segmented. Our method performs much better than a cylinder-based shape primitives method, even when the objects involved (bottles and mugs) are nearly cylindrical.

Another approach to manipulating unknown objects is to estimate the object shape from a recent history of sensor feedback. For example, Dragiev and Toussaint explore an approach that models the implicit shape of an object as a Gaussian process~\cite{Dragiev2011}. Mahler \textit{et al.} do something similar for the purposes of grasping while incorporating tactile feedback~\cite{Mahler2015}. These methods can run into trouble when there is not enough data to fit the implicit shape with high confidence. Both of the above approaches can be viewed as ways of estimating object shape and pose in order to facilitate traditional configuration space planning. The problem of object pose and shape estimation given various amounts of prior data remains an active area of research~\cite{Hinterstoisser2012,Pauwels2015,Wohlhart2015}.

Recently, there has been much advancement in grasping novel objects. Bohg \textit{et al.} provide a survey~\cite{Bohg2014}. Most of these methods are trained in a supervised fashion to predict whether a grasp is stable or not. The present paper can be viewed as extending our prior work in grasp detection~\cite{Gualtieri2016,tenPas2017} to pick-and-place and regrasping.

The approach nearest to ours is by Jiang \textit{et al.} who propose a method for placing new objects in new place areas without explicitly estimating the object's pose~\cite{Jiang2012}. Their placements are sampled instead of, as in our case, fixed. However, they do not jointly reason about the grasp and the place -- the grasp is predefined. This is an important drawback because the type of placement that is desired often has implications on how the grasp should be performed. Their learning method also relies on segmenting the object, segmenting the place area, hand-picked features, and human annotation for place appropriateness.

RL has long been studied for use in robot control. Kober \textit{et al.} survey robotics applications that use RL~\cite{Kober2013}. Since this survey, deep RL has become prominent in robotic manipulation~\cite{Levine2016A,Levine2016B,Viereck2017}. These methods operate on the  motor torque or Cartesian motion command level of the robot controller whereas ours operates at a higher level. 


\section{Problem Statement}
\label{sec:problemStatement}

We consider the problem of grasping, regrasping, and placing a novel object in a desired pose using a robotic arm equipped with a simple gripper. We assume this is a first-order kinematic system such that state is fully described by the geometry of the environment. Also, we assume the agent can act only by executing parameterized reach actions.

The problem can be expressed as an MDP as follows. Let $\Gamma \subseteq \mathbb{R}^3$ denote the portion of work space that is free of obstacles. For simplicity of exposition, suppose that it is known that the free space contains a finite set of $N$ rigid body objects, $O$. Let $\Lambda$ denote a parameter space that describes the space of all possible object shapes. Let $\xi(o) \in \Lambda \times SE(3)$ denote the shape and pose of object $o \in O$. Let $H \in SE(3)$ denote the current pose of the robot hand. The state of the system is fully described by the pose of the hand and the shape and pose of all $N$ objects: $s = (H, \xi(o^1), \dots, \xi(o^N)) \in S = SE(3) \times \{\Lambda \times SE(3)\}^N$. 

We will assume the robot can act only by executing the following parameterized, pre-programmed actions: $\textsc{reach-grasp}(T)$ where $T \in SE(3)$ and $\textsc{reach-place}(t)$ where $t \in \textsc{place} \subset \mathbb{N}$ belongs to a discrete set of pre-programmed reach poses expressed relative to the robot base frame. The total set of available actions is then $A = SE(3) \cup \textsc{place}$.

Given a goal set $G \subset S$, we define a reward function to be $1$ when a goal state is reached and $0$ otherwise. The episode terminates either when a goal state is reached or after a maximum number of actions. Finally, we assume access to a simulator that models the effects of an action $a \in A$ taken from state $s \in S$. For stochastic systems, we assume the simulator draws a sample from the distribution of possible outcomes of an action. Given this formalization of the manipulation problem, we might express it as an MDP $\mathcal{M} = (S,A,\mathcal{T},r)$ with state-action space $S \times A$, unknown but stationary transition dynamics $\mathcal{T}$, and reward function $r(s,a) = 1$ if $s \in G$ and $r(s,a)=0$ otherwise.

A key assumption in this paper is object shape and pose are not observed directly and therefore the MDP defined above is not fully observed. Instead, it is only possible to observe a volumetric occupancy grid, $C(x) \in \{\textsc{occupied}, \textsc{free}, \textsc{unknown}\}$ where $x \in \bar{\Gamma} \subset \Gamma$ is a volumetric grid over $\Gamma$. We assume that $C$ is populated based on depth sensor data obtained during a single time step. (As such, there may be a large number of $\textsc{unknown}$ cells.) Given the above assumptions, the manipulation problem can be expressed as a POMDP $\mathcal{P} = (S,A,\mathcal{T},r,C,\mathcal{O})$, where $\mathcal{O}: S \times A \times C \rightarrow \mathbb{R}$ is the observation probabilities, assumed to be unknown. The goal of this paper is to find policies that maximize the expected sum of discounted rewards over the episode, i.e., reach the goal state in a minimum number of actions.

\section{Approach}
\label{sec:approach}

Solving the POMDP $\mathcal{P}$ using general purpose belief space solvers (e.g.~\cite{Kurniawati2008}) is infeasible because the underlying MDP $\mathcal{M}$ is far too large to solve even if it were fully observed. Instead we propose what we call the \textit{descriptor-based MDP} that encodes $\textsc{reach-grasp}$ actions using a special type of descriptor and encodes belief state implicitly as a short history of states and actions. 

\begin{figure}[ht]
 \vspace{0.1in}
 \centering
 \includegraphics[width=0.750\columnwidth]{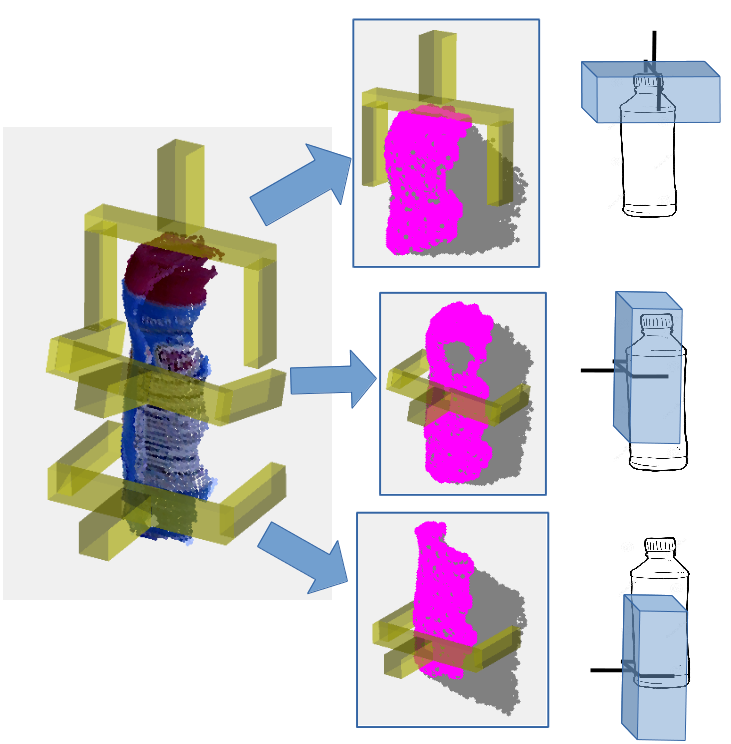}
 \caption{Examples of the grasp descriptor for the three grasps shown on the left. The right column shows the cuboid associated with each grasp. The middle column shows the descriptor -- the visible and occluded points contained within the cuboid.}
 \vspace{-0.2in}
\label{fig:descriptor}
\end{figure}

\subsection{The \textsc{reach-grasp} Descriptor}
\label{sect:grasp_desc}

The $\textsc{reach-grasp}$ descriptor is a key element of our state and action representation, based on the grasp descriptor developed in our prior work~\cite{Gualtieri2016,tenPas2017}. It encodes the relative pose between a robot hand and an object in terms of the portion of the volumetric occupancy grid in the vicinity of a prospective grasp. Let $\mathcal{C} = \{x \in \bar{\Gamma} | C(x) = \textsc{occupied}\}$ denote the voxelized point cloud corresponding to the occupancy grid $C$. Then, the $\textsc{reach-grasp}$ descriptor at pose $T \in SE(3)$ is $D(\mathcal{C},T) = trunc_{\gamma}(T \mathcal{C})$, where $T \mathcal{C}$ is the point cloud in the grasp reference frame, and where $trunc_{\gamma}(X)$ denotes the elements of $X$ that lie within a cuboid centered at the origin with dimensions $\gamma = (\gamma_x, \gamma_y, \gamma_z)$. This is illustrated in Figure~\ref{fig:descriptor}. The middle column shows the $\textsc{reach-grasp}$ descriptors corresponding to the three grasps of the object shown on the left. A $\textsc{reach-grasp}$ descriptor is encoded to the deep network as an image where the points are projected onto planes orthogonal to three different viewing directions and compiled into a single stacked image, $I(D)$, as described in~\cite{Gualtieri2016,tenPas2017}.

\subsection{The Descriptor-Based MDP}
\label{sect:descMDP}

Our key idea is to find goal-reaching solutions to the POMDP $\mathcal{P}$ by reformulating it as an MDP with descriptor-based states and actions. Specifically, we: 1) reparameterize the $\textsc{reach-grasp}$ action using $\textsc{reach-grasp}$ descriptors rather than 6-DOF poses; 2) redefine state as a history of the last two actions visited.

\noindent
\textbf{Action representation:} Recall that the underlying MDP defines two types of actions: $\textsc{reach-grasp}(T)$ where $T$ denotes the pose of the grasp and $\textsc{reach-place}(t)$ where $t \in \textsc{place}$ and $\textsc{place}$ denotes a finite set of place poses. Since RL in a continuous action space can be challenging, we approximate the parameter space of $\textsc{reach-grasp}$ by sampling. That is, we sample a set of $m$ candidate poses for $\textsc{reach-grasp}$: $T_1, \dots, T_m \in SE(3)$. In principle, we can use any sampling method. However, since $\textsc{reach-grasp}$ is intended to reach toward \textit{grasp} configurations, we use grasp pose detection (GPD)~\cite{Gualtieri2016,tenPas2017} to generate the samples. Each of the candidate poses generated by GPD is predicted to be a pose from which closing the robot hand is expected to result in a grasp (although the grasp could be of any object).

Since we are sampling candidate parameters for $\textsc{reach-grasp}$, we need a way to encode these choices to the action-value function. Normally, in RL, the agent has access to a fixed set of action choices where each choice always results in the same distribution of outcomes. However, since we are now sampling actions, this is no longer the case, and we need to encode actions to the action-value function differently. In this paper, we encode each target pose candidate for $\textsc{reach-grasp}$ by the corresponding \textsc{reach-grasp} descriptor, $D_i = D(\mathcal{C},T_i), i \in [1,m]$. Essentially, the descriptor encodes each target pose candidate by the image describing what the point cloud nearby the target pose looks like. The total action set consists of the set of descriptors corresponding to sampled reach-grasps, $\textsc{reach-grasp}(D(\mathcal{C},T_i)), i \in [1,m]$, and the discrete set of reach-places adopted from the underlying POMDP, $\textsc{reach-place}(i), i \in \textsc{place}$: $A = [1,m] \cup \textsc{place}$.

\begin{figure}[ht]
 \centering
 \includegraphics[width=0.750\columnwidth]{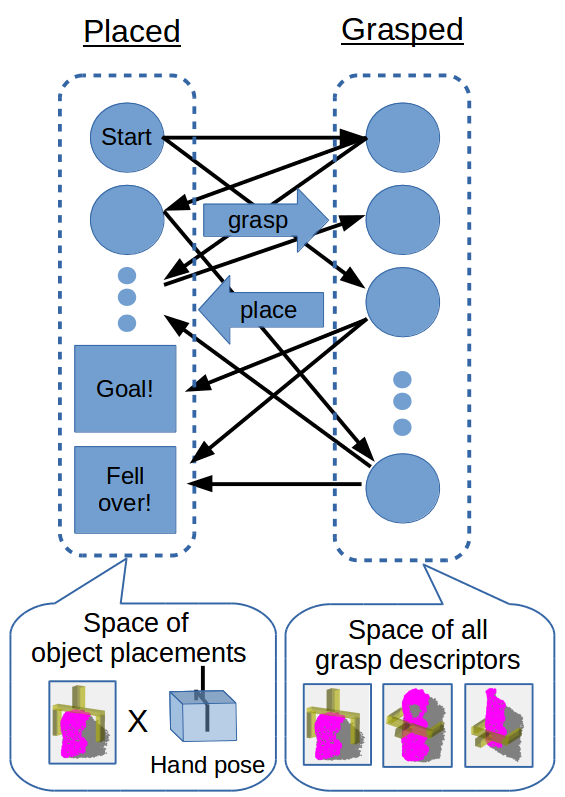}
 \caption{The descriptor-based MDP. States on the right are those where an object is grasped. All other states are on the left.}
 \vspace{-0.1in}
\label{fig:mdp}
\end{figure}

\noindent
\textbf{State representation:} We encode state as the history of the $M$ most recent reach actions where $\textsc{reach-grasp}$ actions are represented using the corresponding descriptors. In all of our experiments, $M \leq 2$. Figure~\ref{fig:mdp} illustrates the resulting state-action space. The set of blue circles on the right labeled ``Space of all grasp descriptors'' denotes the set of states where an object has been grasped. This is a continuous space of states equal to the set of $\textsc{reach-grasp}$ descriptors resulting from the most recent $\textsc{reach-grasp}$ action, $\{trunc_\gamma(\mathcal{C}) | \mathcal{C} \subset \bar{\Gamma} \}$. The set of blue circles on the left labeled ``Space of object placements'' represents the set of states where an object has been placed somewhere in the environment. These states are encoded as the history of the two most recent reach actions: the $\textsc{reach-place}$ action taken on the last time step and the descriptor that encodes the $\textsc{reach-grasp}$ action taken two time steps ago. All together, a state in this new MDP is a point in $S = \{trunc_\gamma(\mathcal{C}) | \mathcal{C} \subset \bar{\Gamma} \} \times \textsc{place}$. The state labeled ``Goal!'' in Figure~\ref{fig:mdp} denotes an absorbing state where the object has been placed correctly, and the state labeled ``Fell over!'' denotes an absorbing state where the object has been placed incorrectly. When the agent reaches either of these states, it obtains a final reward and the episode ends.

\noindent
\textbf{Reward:} Our agent obtains a reward of $+1$ when it reaches a placement state that satisfies the desired problem constraints, and otherwise, it obtains zero reward.

\subsection{The Simulator}

Deep RL requires such an enormous amount of experience that it is difficult to learn control policies on real robotic hardware without spending months or years in training~\cite{Levine2016B,Levine2016A}. As a result, learning in simulation is basically a requirement. Fortunately, our formulation of the manipulation problem in terms of pre-programmed, parameterized actions simplifies the simulations. Instead of needing to simulate \textit{arbitrary} contact interactions, we only need a mechanism for simulating the grasp that results from executing $\textsc{reach-grasp}(T)$ and the object placement that results from executing $\textsc{reach-place}(t)$. The former can be simulated by evaluating standard grasp quality metrics~\cite{Murray1994}. The later can be simulated by evaluating sufficient conditions to determine whether an object will fall over given the executed placement. Both are easy to evaluate in OpenRAVE~\cite{Diankov2010}, the simulator used in this work.

\subsection{The Action-Value Function}

\begin{figure}[ht]
 \centering
 \includegraphics[width=0.750\columnwidth]{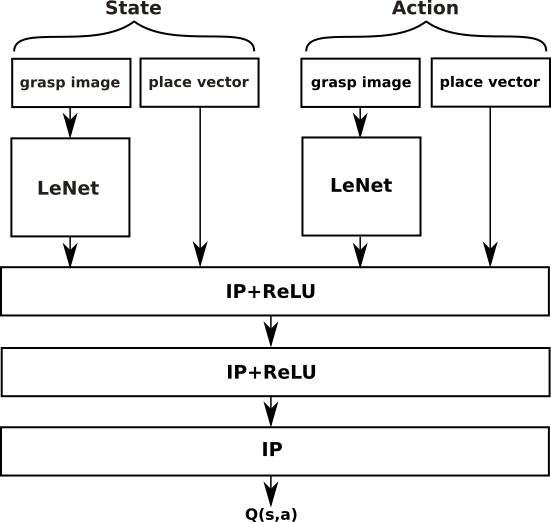}
 \caption{Convolutional neural network architecture used to encode the action-value function, i.e., the Q-function.}
 \vspace{-0.1in}
\label{fig:cnn}
\end{figure}

We approximate the action-value function using the convolutional neural network (CNN) shown in Figure~\ref{fig:cnn}. The input is an encoding of the state and the action, and the output is a scalar, real value representing the value of that state-action pair. This structure is slightly different than that used in DQN~\cite{Mnih2015} where the network has a number of outputs equal to the number of actions. Here, the fact that our MDP uses sampled reach actions means that we must take action as an input to the network. The action component of the input is comprised of the $\textsc{reach-grasp}$ descriptor (encoded as a $60 \times 60 \times 12$ stacked image as described in Section~\ref{sect:grasp_desc}) denoting the $\textsc{reach-grasp}$ parameter and a one-hot vector denoting the $\textsc{reach-place}$ parameter. When the agent selects $\textsc{reach-grasp}$, the grasp descriptor is populated and the place vector is set to zero. When a $\textsc{reach-place}$ is selected, the grasp descriptor is set to zero and the place vector is populated. 

The state component of the input is also comprised of a $\textsc{reach-grasp}$ descriptor and a place vector. However, here these two parameters encode the recent history of actions taken (Section~\ref{sect:descMDP}). After executing a grasp action, the grasp descriptor component of state is set to a stored version of the descriptor of the selected grasp and the place vector is set to zero. After executing a place action, the grasp descriptor retains the selected grasp and the place component is set to the just-executed place command, thereby implicitly encoding the resulting pose of the object following the place action. Each grasp image (both in the action input and the state input) is processed by a CNN similar to LeNet~\cite{LeCun1998}, except the output has 100 hidden nodes instead of 10. These outputs, together with the place information, are then concatenated and passed into two 60-unit fully connected, inner product (IP) layers, each followed by rectifier linear units (ReLU). After this there is one more inner product to produce the scalar output. 

\subsection{Learning Algorithm}

Our learning algorithm is shown in Algorithm~\ref{alg:batchSarsa}. This is similar to standard DQN~\cite{Mnih2015} with a couple of differences. First, we use a variant of Sarsa~\cite{Rummery1994} rather than Q-learning because the large action branching factor makes the $\max_{a \in A} Q(s,a)$ in Q-learning expensive to evaluate and because Sarsa is known to perform slightly better on non-Markov problems. Second, we do not run a single stochastic gradient descent (SGD) step after each experience. Instead, we collect \textit{nEpisodes} of experience before labeling the experience replay database using the most recent neural network weights. Every \textit{nEpisodes} additional experiences, we run \textit{nIterations} of SGD using Caffe~\cite{Jia2014}. For the experiments in this paper, the learning algorithm is run only in simulation; although it could be used to fine-tune the network weights on the actual hardware.

\begin{figure}[th]
\vspace{0.07in}
\removelatexerror
\begin{algorithm}[H]
\DontPrintSemicolon
\caption{Sarsa implementation for pick and place}
\label{alg:batchSarsa}
\SetKwFunction{Pick}{Pick}
\SetKwFunction{Place}{Place}
\SetKwFunction{Subsample}{Subsample}
\For {$i \gets 1:$ nTrainingRounds} {
  \For {$j \gets 1:$ nEpisodes} {
    Choose random object(s) from training set\;
    Place object(s) in a random configuration\;
    Sense point cloud $\mathcal{C}$ and detect grasps $\mathcal{G}$\;
    $s \gets$ initial state \;
    $a \gets \Pick(.)$ ($\epsilon$-greedy)\;
    \For {$t \gets 1 : \textit{maxTime}$} {
      $(r, s') \gets \mathcal{T}(s,a)$ \;
      \If {$a = \Pick(.)$} {
        $a' \gets \Place(.)$ ($\epsilon$-greedy) \;
     }
     \ElseIf {$a = \Place(p)|p \in P_{\textit{temp}}$} {
        Sense point cloud $\mathcal{C}$ and detect grasps $\mathcal{G}$\;
       $a' \gets \Pick(.)$  ($\epsilon$-greedy) \;
     }
     \ElseIf {$a = \Place(p)|p \in P_{\textit{final}}$} {
       $a' \gets \textit{null}$
     }
     Add $(s,a,r,s',a')$ to database \;
     \lIf {$s'$ is terminal} {break}
     $a \gets a'$; $s \gets s'$ \;
    }
  }
  Prune database if it is larger than \textit{maxExperiences} \;
  Label each database entry $(s,a)$ with $r + \gamma Q(s', a')$ \;
  Run Caffe for \textit{nIterations} on database \;
}
\label{alg:training}
\end{algorithm}
\vspace{-0.2in}
\end{figure}

\section{Experiments in Simulation}
\label{sec:simulationResults}

We performed experiments in simulation to evaluate how well our approach performs on pick-place and regrasping problems with novel objects. To do so, we obtained objects belonging to two different categories for experimentation: a set of 73 bottles and a set of 75 mugs -- both in the form of mesh models from 3DNet~\cite{Wohlkinger2012}. Both object sets were partitioned into a 75\%/25\% train/test split.

\subsection{Experimental Scenarios}

There were three different experimental scenarios, \textit{two-step-isolation}, \textit{two-step-clutter}, and \textit{multi-step-isolation}. In \textit{two-step-isolation}, an object was selected at random from the training set and placed in a random pose in isolation on a tabletop. The goal condition was a right-side-up placement in a particular position on a table. In this scenario, the agent was only allowed to execute one grasp action followed by one place action (hence the ``two-step'' label). \textit{Two-step-clutter} was the same as \textit{two-step-isolation} except a set of seven objects was selected at random from the same object category and placed in random poses on a tabletop as if they had been physically dumped onto the table.

The \textit{multi-step-isolation} scenario was like \textit{two-step-isolation} except multiple picks/places were allowed for up to 10 actions (i.e., \textit{maxTime}=10). Also, the goal condition was more restricted: the object needed to be placed upright, inside of a box rather than on a tabletop. Because the target pose was in a box, it became impossible to successfully reach it without grasping the object from the top before performing the final place (see Figure~\ref{fig:robotExperimentIllustration}, bottom). Because the object could not always be grasped in the desired way initially, this additional constraint on the goal state sometimes forced the system to perform a regrasp in order to achieve the desired pose.

In all scenarios, point clouds were registered composites of two clouds taken from views above the object and $90^\circ$ apart: a single point cloud performs worse, presumably because features relevant for determining object pose are unobserved. In simulation, we assumed picks always succeed, because the grasp detector was already trained to recognize stable grasps with high probability~\cite{Gualtieri2016,tenPas2017}~\footnote{It is possible to train grasping from the same reward signal, but this would require longer simulations. Empirically, this assumption did not lead to many grasp failures on the real robot (see Section~\ref{sect:real-robot}).}. A place was considered successful only if the object was placed within 3 cm of the table and 20 degrees of the vertical in the desired pose.

\subsection{Algorithm Variations}

The algorithm was parameterized as follows. We used $70$ training rounds ($\textit{nTrainingRounds}=70$ in Algorithm~\ref{alg:batchSarsa}) for the two-step scenarios and $150$ for the multi-step scenario. We used $1,000$ episodes per training round (\textit{nEpisodes} $=1,000$). For each training round we updated the CNN with $5,000$ iterations of SGD with a batch size of 32. \textit{maxExperiences} was $25,000$ for the two-step scenarios and $50,000$ for the multi-step scenario. For each episode, bottles were randomly scaled in height between 10 and 20 cm. Mugs were randomly scaled in height between 6 and 12 cm. We linearly decreased the exploration factor $\epsilon$ from 100\% down to 10\% over the first $18$ training rounds.


We compared the performance of Algorithm~\ref{alg:batchSarsa} on two different types of \textsc{reach-grasp} descriptors. In the \textit{standard} variation, we used descriptors of the standard size ($10 \times 10 \times 20$ cm). In the \textit{large-volume} (LV) variation, we used descriptors evaluated over a larger volume ($20 \times 20 \times 40$ cm) but with the same image resolution.

We also compared with two baselines. The first was the \textit{random baseline}, where grasp and place actions were chosen uniformly at random. The second was the \textit{shape primitives baseline}, where object pose was approximated by segmenting the point cloud and fitting a cylinder. Although it is generally challenging to fit a shape when the precise geometry of the object to be grasped is unknown, we hypothesized that it could be possible to obtain good pick-place success rates by fitting a cylinder and using simple heuristics to decide which end should be up. We implemented this as follows. First, we segment the scene into k clusters, using k-means ($k=1$ for isolation and $k=7$ for clutter). Then we fit a cylinder to the most isolated cluster using MLESAC~\cite{Torr2000}. We select the grasp most closely aligned with and nearest to the center of the fitted cylinder. The height of the placement action is determined based on the length of the fitted cylinder. The grasp up direction is chosen to be aligned with the cylinder half which contains fewer points. In order to get the shape primitive baseline to work, we had to remove points on the table plane from the point cloud. Although our learning methods do not require this and work nearly as well either way, we removed the table plane in all simulation experiments for consistency.

\subsection{Results for the Two-Step Scenarios}

\begin{figure*}[t]
  \vspace{0.1in}
  \centering
  \includegraphics[height=1.9in,trim={0.25in 0in 0.3in 0in},clip]{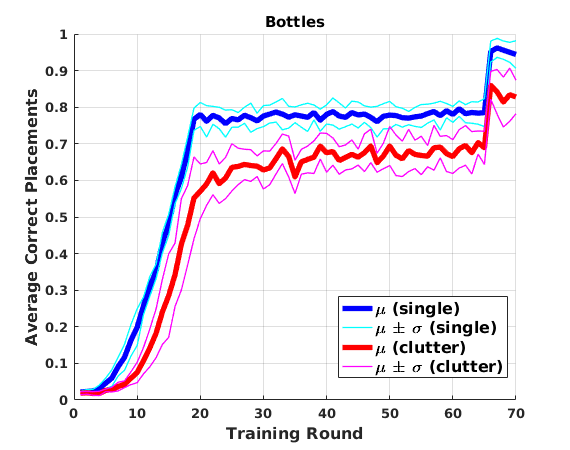}
  \includegraphics[height=1.9in,trim={0.25in 0in 0.3in 0in},clip]{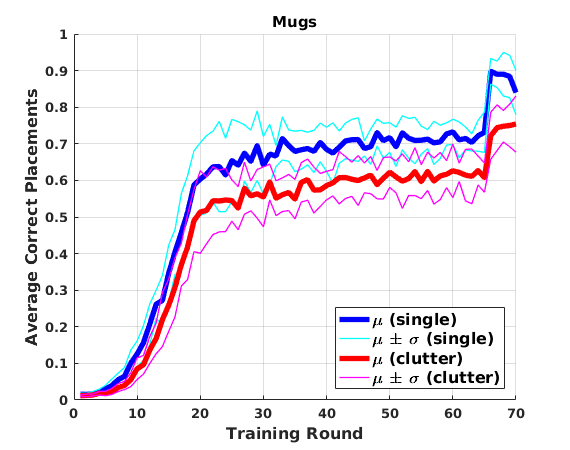}
  \includegraphics[height=1.9in,trim={0.25in 0in 0.3in 0in},clip]{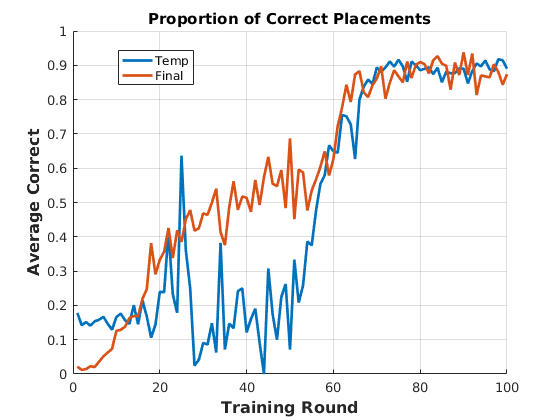}
  \caption{\textbf{Left} and \textbf{center}. Average of 10 learning curves for the two-step scenario. The ``training round'' on the horizontal axis denotes the number of times Caffe had been called for a round of $5,000$ SGD iterations. The left plot is for bottles and the center for mugs. Blue denotes single objects and red denotes clutter. Curves for mean plus and minus standard deviation are shown in lighter colors. The sharp increase in performance during the last five rounds in each graph is caused by dropping the exploration factor ($\epsilon$) from 10\% to 0\% during these rounds. \textbf{Right.} One multi-step realization with mugs in isolation. Red line: number of successful pick-place trials as a function of training round. Blue line: number of successful non-goal placements executed.}
  \label{fig:learningCurves}
  \vspace{-0.2in}
\end{figure*}

Figure~\ref{fig:learningCurves} shows learning curves for the two-step-isolation and two-step-clutter contingencies for bottles (left) and mugs (center) averaged over 10 runs. Table~\ref{tab:twoStepResults} shows place success rates when the test objects were used.

Several results are worth highlighting. First, our algorithm does very well with respect to the baselines. The random baseline (last row in Table~\ref{tab:twoStepResults}) succeeds only 2\% of the time -- suggesting that the problem is indeed challenging. The shape primitives baseline (where we localize objects by fitting cylinders) also does relatively poorly: it succeeds at most only 43\% of the time for bottles and only 12\% of the time for mugs. Second, place success rates are lower when objects are presented in clutter compared to isolation: 100\% success versus 87\% success rates for bottles; 84\% versus 75\% success for mugs. Also, if evaluation is to be in clutter (resp. isolation), then it helps to train in clutter (resp. isolation) as well: if trained only in isolation, then clutter success rates for bottles drops from 87\% to 67\%; clutter success rates for mugs drops from 75\% to 60\%. Also, using the LV descriptor can improve success rates in isolation (an increase of 84\% to 91\% for mugs), but hurts when evaluated in clutter: a decrease from 87\% to 80\% for bottles; a decrease from 75\% to 70\% for mugs. We suspect that this drop in performance reflects the fact that in clutter, the large receptive field of the LV descriptor encompasses ``distracting'' information created by other objects nearby the target object (remember we do not use segmentation) \cite{Mnih2014}.

\begin{table}[t]
  \vspace{0.1in}
  \centering
  \begin{tabular}{|c | c | c |} 
    \hline
    \textsubscript{Trained With} / \textsuperscript{Tested With} & Bottle in Iso. & Bottles in Clut. \\
    \hline
    Isolation & 1.00 & 0.67\\
    \hline
    Clutter & 0.78 & 0.87\\
    \hline
    Isolation LV & 0.99 & 0.47\\
    \hline
    Clutter LV & 0.96 & 0.80\\
    \hline
    Shape Primitives Baseline & 0.43 & 0.24\\
    \hline
    Random Baseline & 0.02 & 0.02\\
    \hline
    \hline
    \textsubscript{Trained With} / \textsuperscript{Tested With} & Mug in Iso. & Mugs in Clut. \\
    \hline
    Isolation & 0.84 & 0.60 \\
    \hline
    Clutter & 0.74 & 0.75 \\
    \hline
    Isolation LV & 0.91 & 0.40 \\
    \hline
    Clutter LV & 0.67 & 0.70\\
    \hline
    Shape Primitives Baseline & 0.08 & 0.12\\
    \hline
    Random Baseline & 0.02 & 0.02\\
    \hline
  \end{tabular}
  \caption{Average correct placements over 300 episodes for bottles (top) and mugs (bottom) using test set, after training.}
  \label{tab:twoStepResults}
  \vspace{-0.2in}
\end{table}

\subsection{Results for the Multi-Step Scenario}

Training for the multi-step-isolation scenario is the same as it was in the two-step scenario except we increased the number of training rounds because the longer policies took longer to learn. We only performed this experiment using mugs (not bottles) because it was difficult for our system to grasp many of the bottles in our dataset from the top. Figure~\ref{fig:learningCurves} shows the number of successful non-goal and goal placements as a function of training round~\footnote{Non-goal placements were considered successful if the object was 3 cm or less above the table. Any orientation was allowed. Unsuccessful non-goal placements terminate the episode.}. Initially, the system does not make much use of its ability to perform intermediate placements in order to achieve the desired goal placement, i.e., to pick up the mug, put it down, and then pick it up a second time in a different way. This is evidenced by the low values for non-goal placements (the blue line) prior to round 60. However, after round 60, the system learns the value of the non-goal placement, thereby enabling it to increase its final placement success rate to is maximum value (around 90\%). Essentially, the agent learns to perform a non-goal placement when the mug cannot immediately be grasped from the top or if the orientation of the mug cannot be determined from the sensor perception. After learning is complete, we obtain an 84\% pick and place success rate averaged over 300 test set trials.


\section{Experiments on a Real Robot}
\label{sect:real-robot}

We evaluated the same three scenarios on a real robot: \textit{two-step-isolation}, \textit{two-step-clutter}, and \textit{multi-step-isolation}. As before, the two step scenarios were evaluated for both bottles and mugs, and the multi-step scenario was evaluated for only mugs. All training was done in simulation, and fixed CNN weights were used on the real robot.

\begin{figure}[th]
  \vspace{0.1in}
  \centering
  \includegraphics[height=0.71in,trim={0in 0.5in 0in 0.1in},clip]{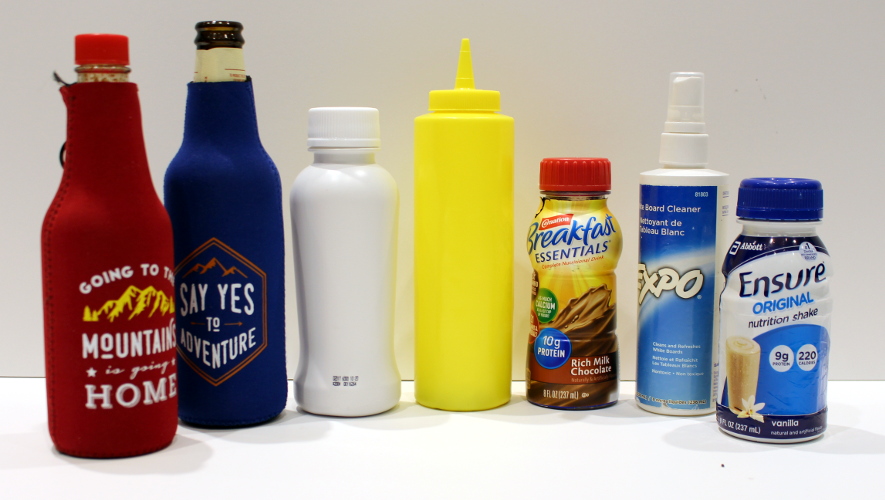}
  \includegraphics[height=0.71in,trim={0in 0.5in 0in 0.2in},clip]{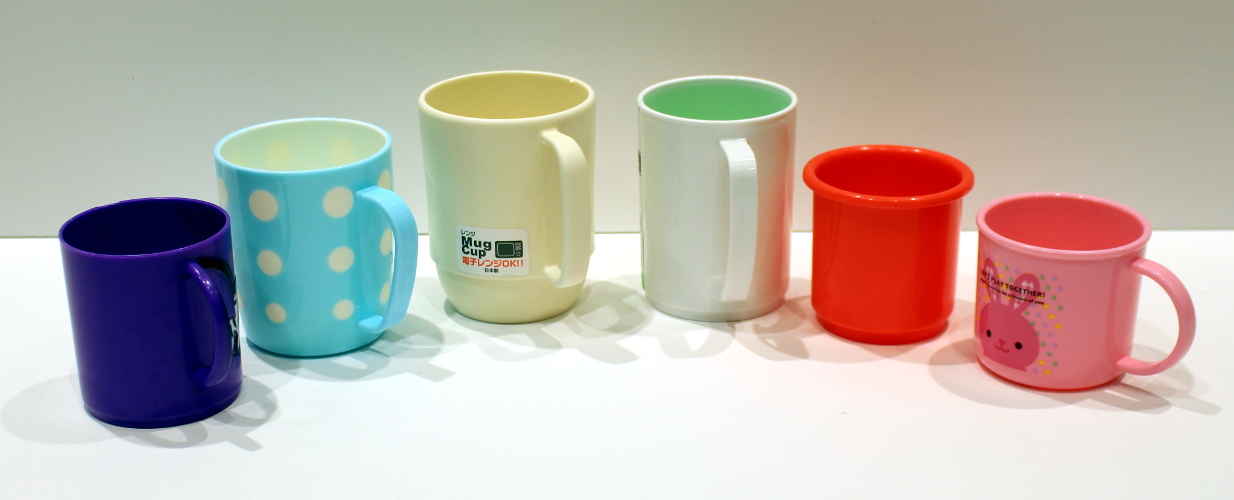}
  \caption{The seven novel bottles and six novel mugs used to evaluate our approach in the robot experiments.}
  \label{fig:bottlesmugs}
  \vspace{-0.2in}
\end{figure}

\begin{figure*}[t]
  \vspace{0.1in}
  \centering
  \includegraphics[height=0.9in,trim={2in 0in 7in 3in},clip]{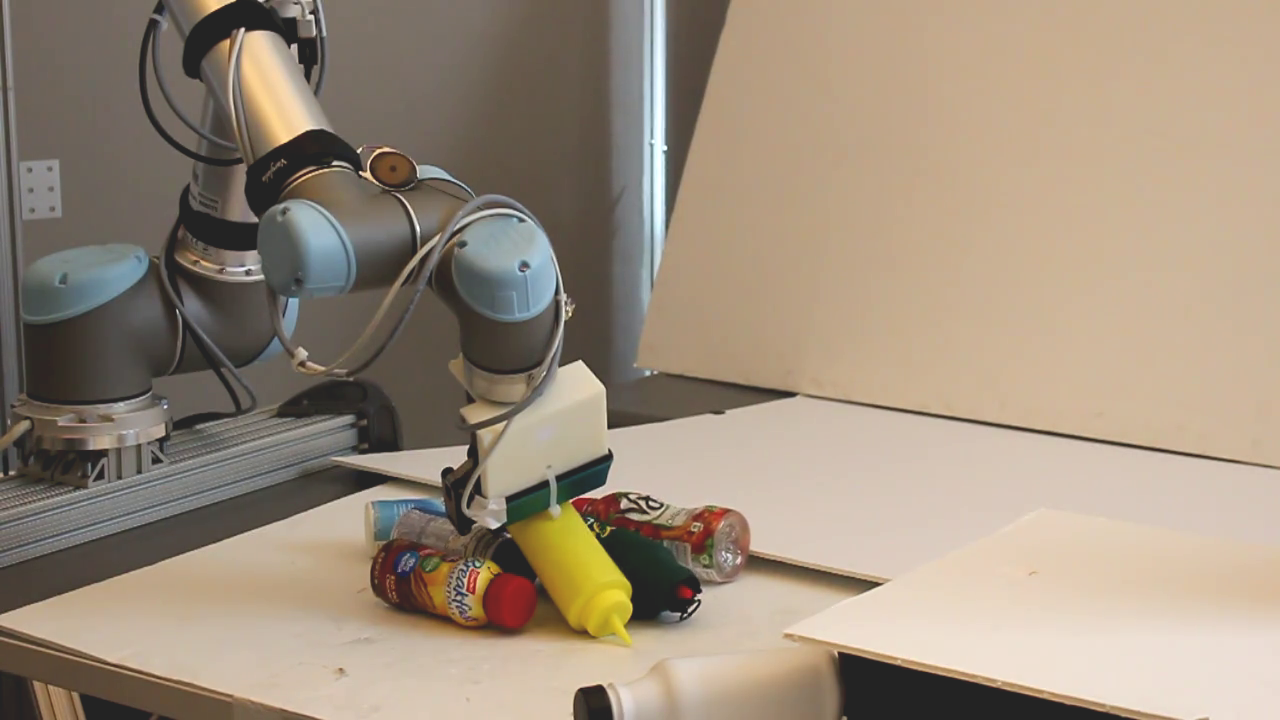}
  \includegraphics[height=0.9in,trim={7in 0in 2in 3in},clip]{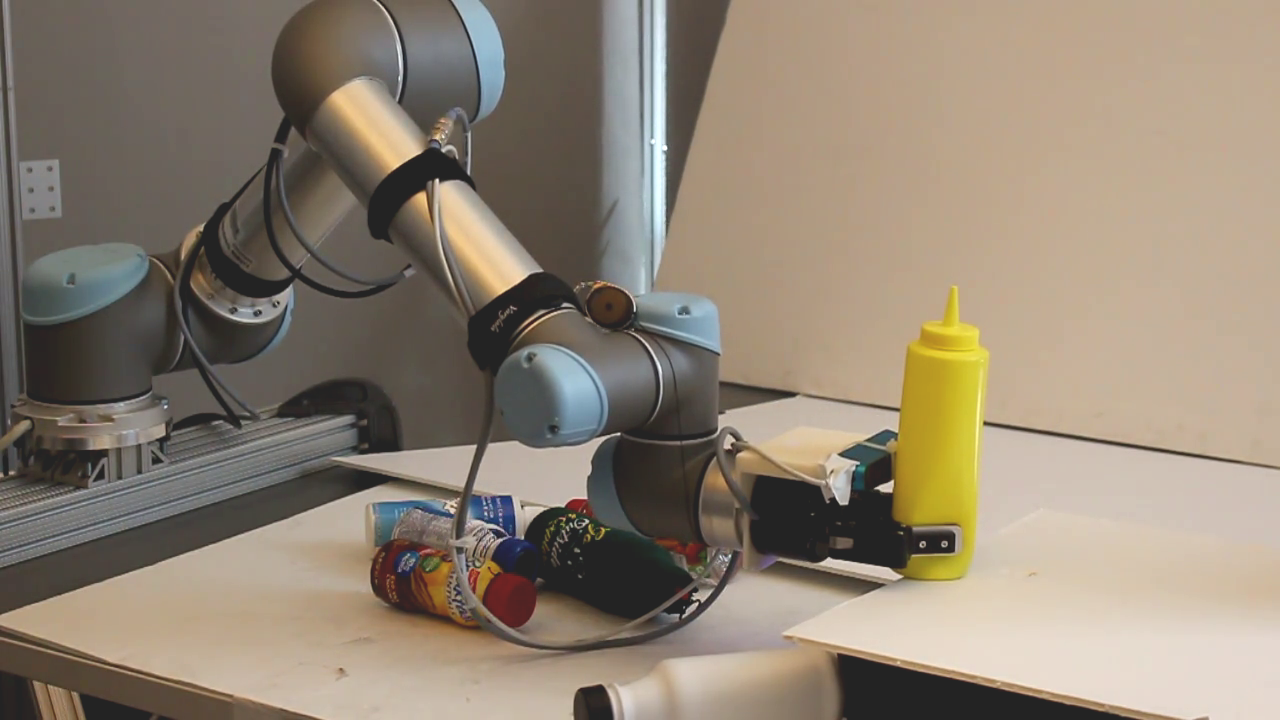}
  \includegraphics[height=0.9in,trim={2in 0in 7in 3in},clip]{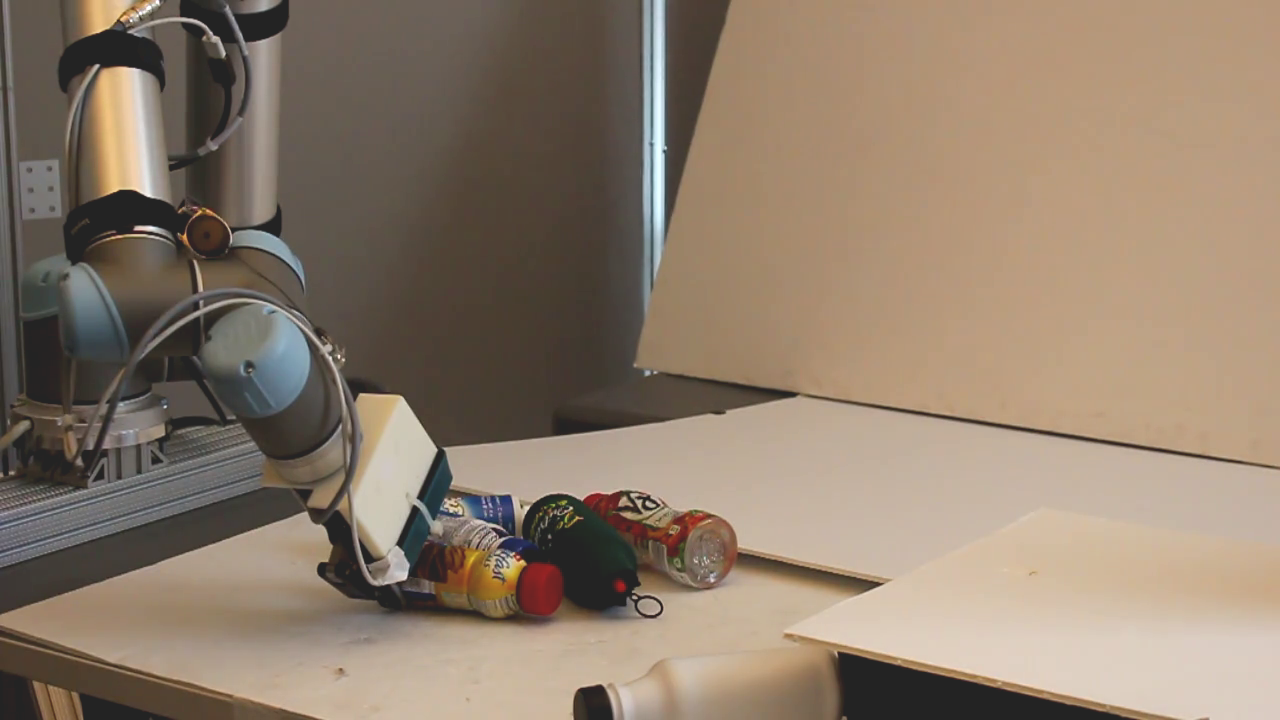}
  \includegraphics[height=0.9in,trim={7in 0in 2in 3in},clip]{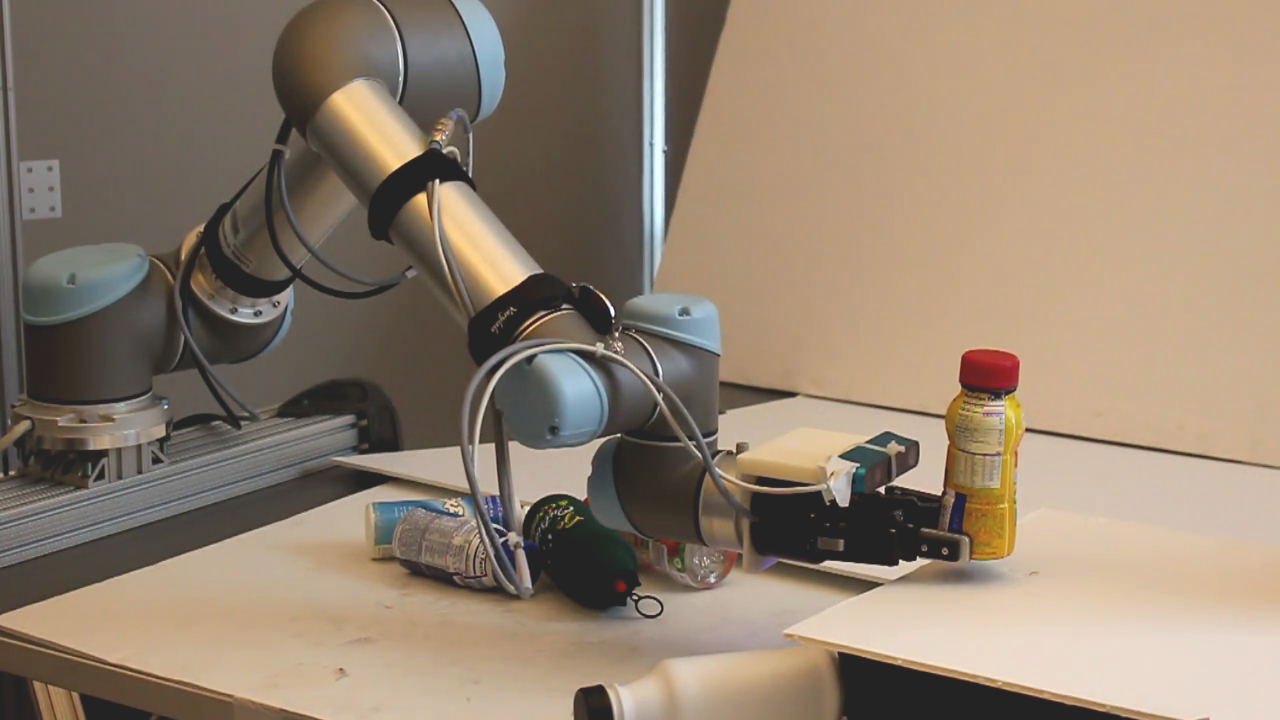}
  \includegraphics[height=0.9in,trim={2in 0in 7in 3in},clip]{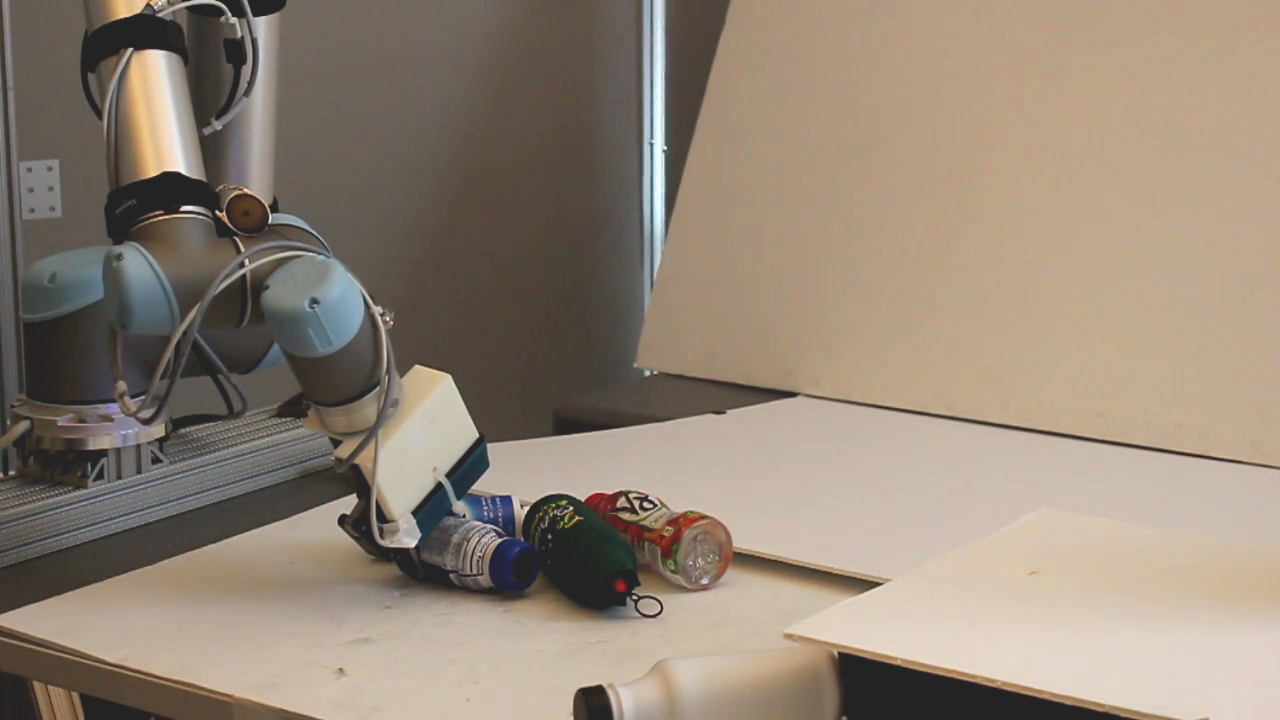}
  \includegraphics[height=0.9in,trim={7in 0in 2in 3in},clip]{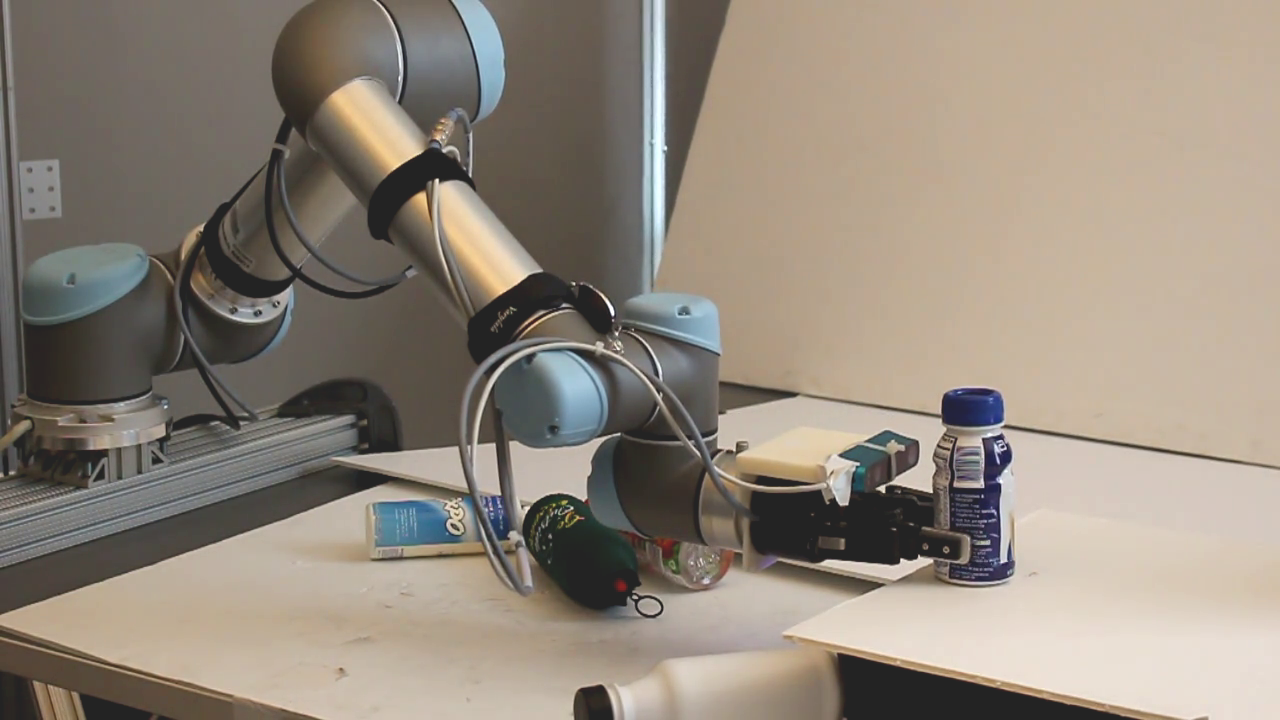}\\
  \includegraphics[height=0.9in,trim={3.05in 2in 3in 1in},clip]{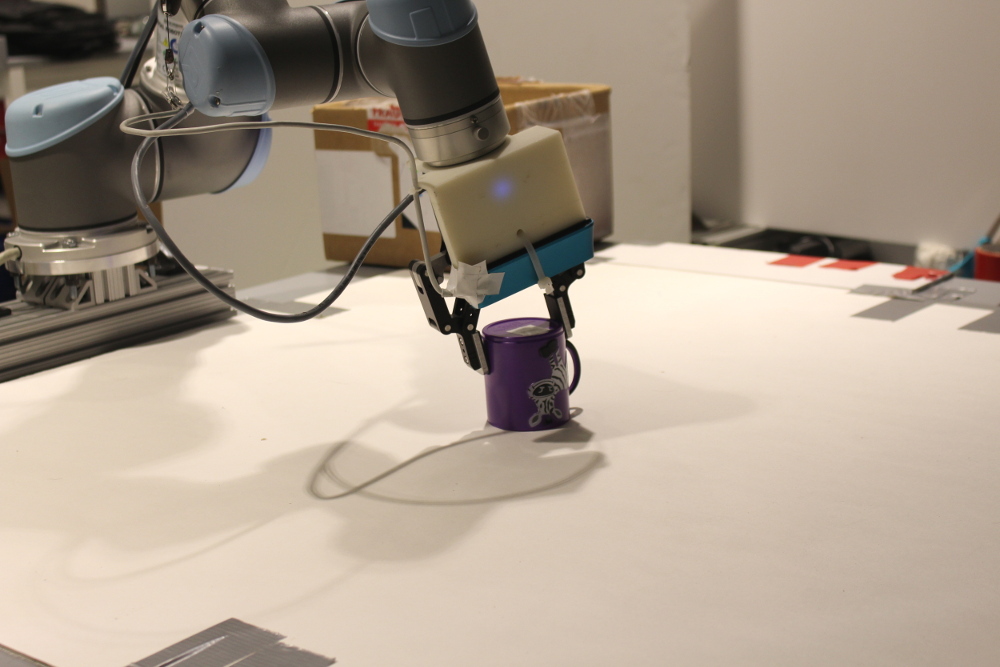}
  \includegraphics[height=0.9in,trim={3.05in 2in 3in 1in},clip]{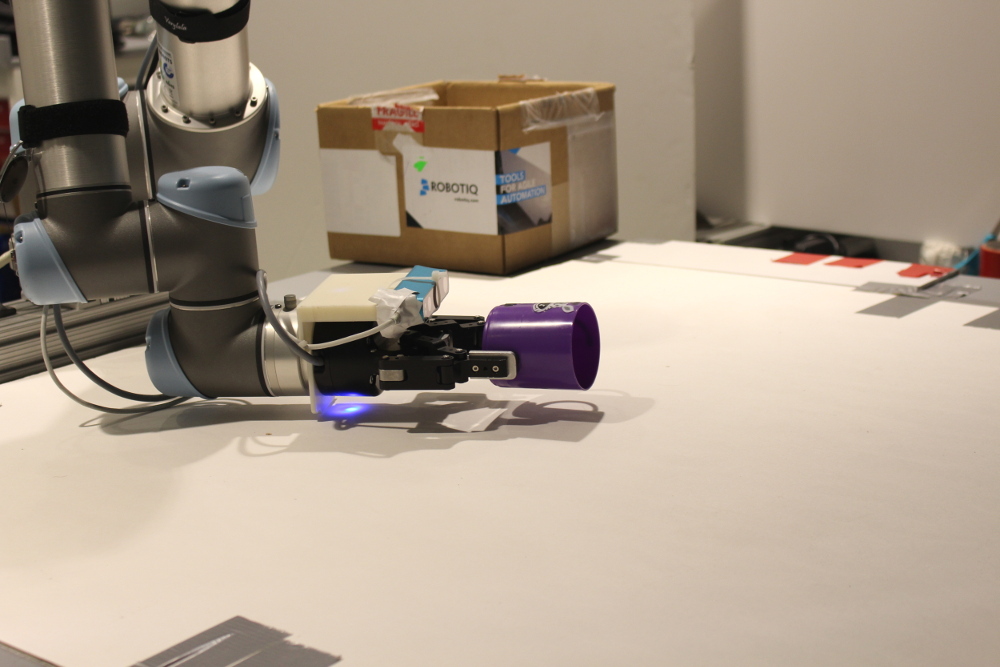}
  \includegraphics[height=0.9in,trim={3.05in 2in 3in 1in},clip]{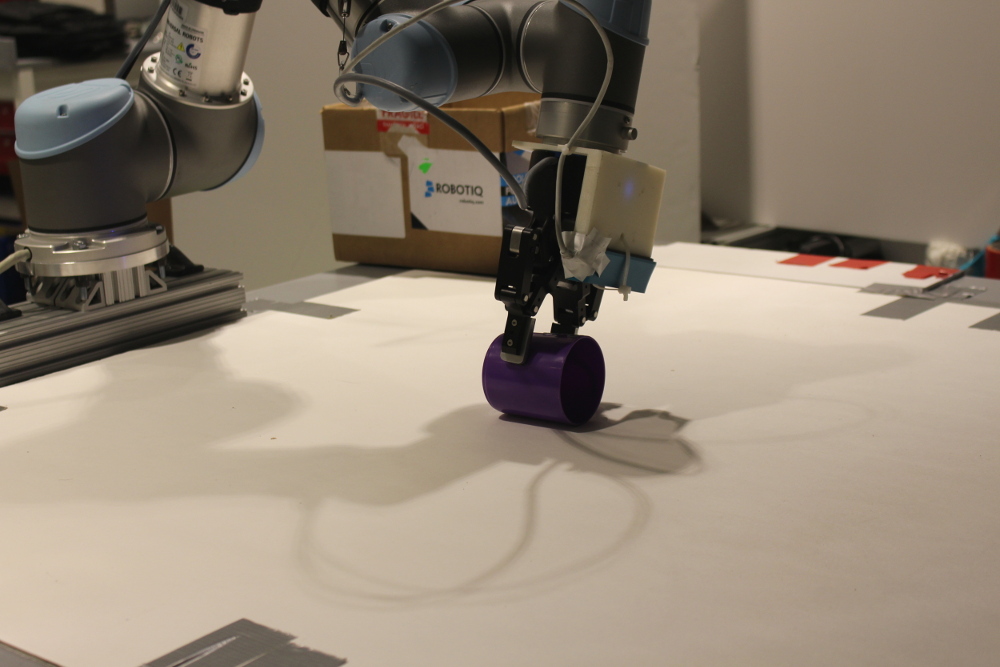}
  \includegraphics[height=0.9in,trim={3.05in 2in 3in 1in},clip]{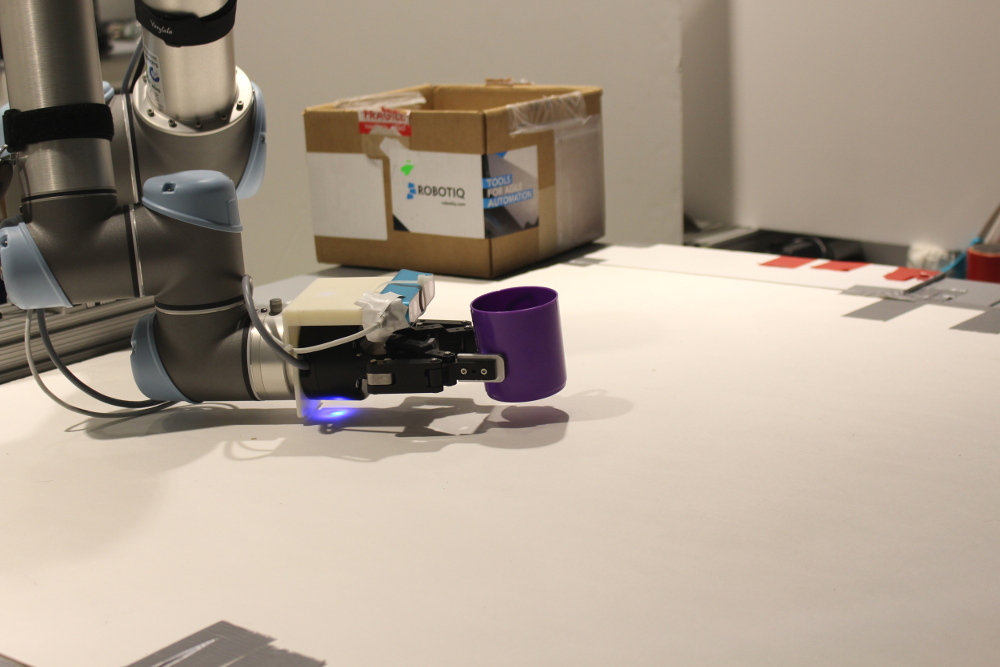}
  \includegraphics[height=0.9in,trim={3.05in 2in 3in 1in},clip]{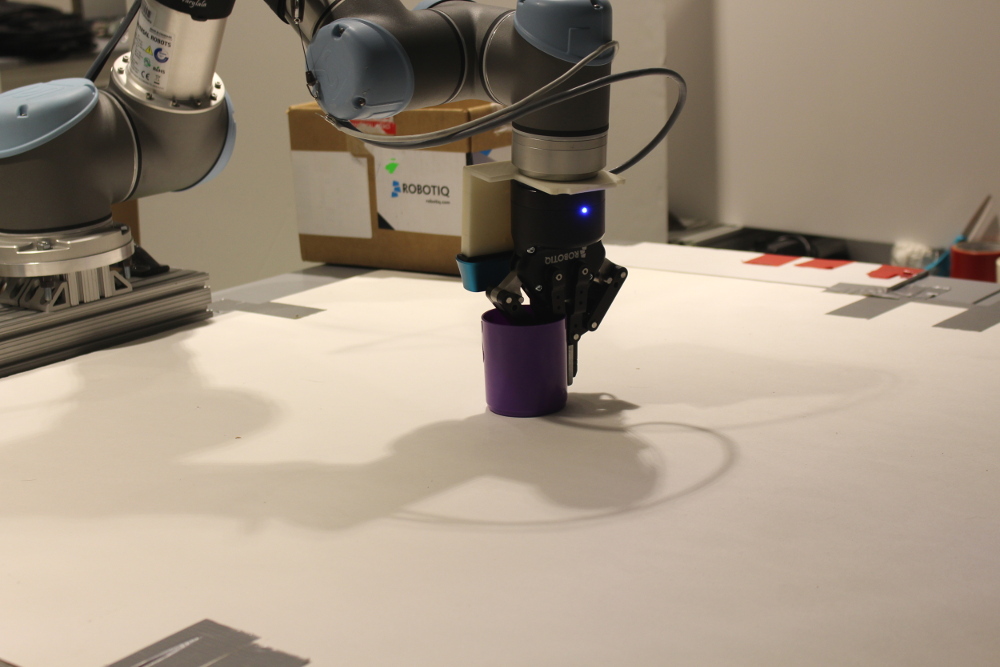}
  \includegraphics[height=0.9in,trim={3.05in 2in 3in 1in},clip]{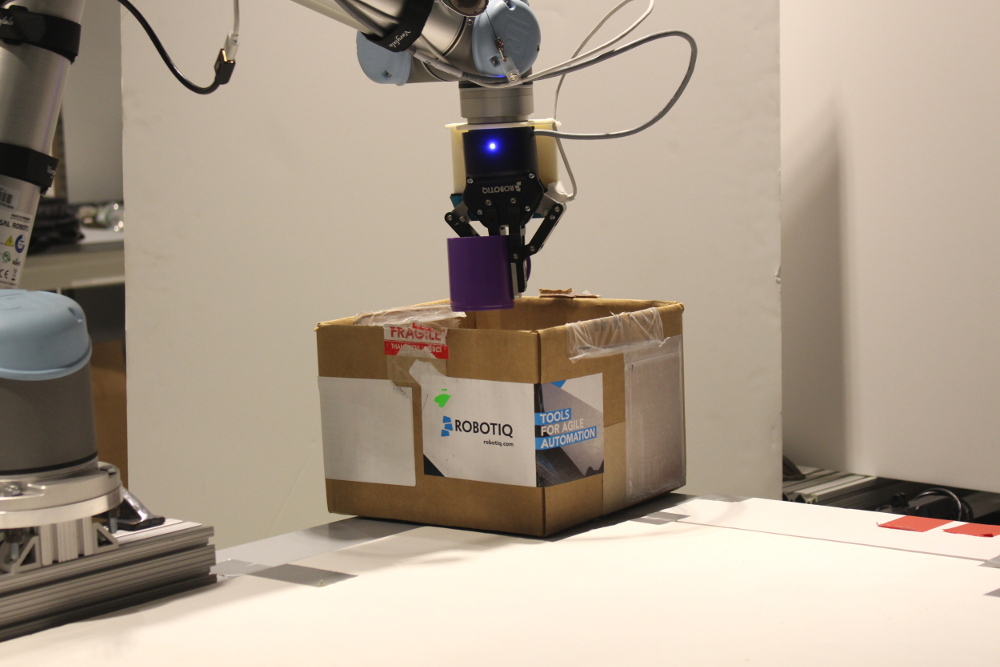}
  \caption{\textbf{Top}. Two-step-clutter scenario for bottles. First three objects are placed right-side-up and without falling over. \textbf{Bottom}. Multi-step-isolation scenario for a mug. The mug is initially upside-down, so must be flipped around before it can be put upright into the box.}
  \label{fig:robotExperimentIllustration}
  \vspace{-0.2in}
 \end{figure*}

The experiments were performed by a UR5 robot with 6 DOFs, equipped with a Robotiq parallel-jaw gripper and a wrist-mounted Structure depth sensor (Figure~\ref{fig:robotExperimentIllustration}). Two sensor views were always taken from fixed poses, $90^\circ$ apart. The object set included 7 bottles and 6 mugs, as shown in Figure~\ref{fig:bottlesmugs}. We used only objects that fit into the gripper, would not shatter when dropped, and had a non-reflective surface visible to our depth sensor. Some of the lighter bottles were partially filled so small disturbances (e.g., sticking to fingers) would not cause a failure. Figure~\ref{fig:robotExperimentIllustration} shows several examples of our two-step scenario for bottles presented in clutter.


Unlike in simulation, the UR5 requires an IK solution and motion plan for any grasp or place pose it plans to reach to. For grasps, GPD returns many grasp choices. We sort these by their pick-place Q-values in descending order and select the first reachable grasp. For places, the horizontal position on the shelf and orientation about the vertical (gravity) axis do not affect object uprightness or the height of the object. Thus, these variables were chosen to suit reachability.

After testing some trials on the UR5, we found we needed to adjust a couple of training/simulation parameters. First, we changed the conditions for a successful place in simulation because, during our initial experiments, we found the policy sometimes selected placements that caused the objects to fall over. As a result, we adjusted the maximum place height in simulation from 3 cm to 2 cm and changed the reward function to fall off exponentially from $+1$ for altitudes higher than 2 cm. Second, we raised the acceptance threshold~\footnote{GPD outputs a machine-learned probability of a stable (i.e., force closure) grasp. The \textit{threshold} is the grasp stability probability above which grasps are accepted.} used by our grasp detector, GPD~\cite{Gualtieri2016,tenPas2017}.


\begin{table}[th]
  \centering
  \begin{tabular}{|c | c | c |c | c | c|}
    \hline
             & 1 Bottle & 7 Bottles & 1 Mug & 6 Mugs & Regrasp\\
    \hline
    Grasp           & 0.99 & 0.97 & 0.96 & 0.93 & 0.94\\
    \hline
    FinalPlace      & 0.98 & 0.94 & 0.93 & 0.87 & 1.00\\
    \hline
    TempPlace       & -    & -    & -    & -    & 1.00\\
    \hline
    EntireTask      & 0.97 & 0.92 & 0.90 & 0.80 & 0.68\\
    \hline
    \hline
    $n$ Trials      & 112  & 107  & 96   & 96   & 72\\
    \hline
    \hline
    UpsideDown       & 0    & 4    & 5    & 10   & 0 \\
    \hline
    Sideways        & 0    & 0    & 0    & 2    & 0 \\
    \hline
    FellOver        & 2    & 2    & 1    & 0    & 0 \\    
    \hline
    $t>10$          & -    & -    & -    & -    & 12\\
    \hline
  \end{tabular}
  \caption{(Top) Success rates for grasp, temporary place, final place, and entire task. (Bottom) Placement error counts by type. Results are averaged over the number of trials (middle).}
  \label{tab:robotResultsSummary}
  \vspace{-0.2in}
\end{table}

Table~\ref{tab:robotResultsSummary} summarizes the results from our robot experiments. We performed 483 pick and place trials over five different scenarios. Column one of Table~\ref{tab:robotResultsSummary} shows results for pick and place for a single bottle presented in isolation averaged over all bottles in the seven-bottle set. Out of 112 trials, 99\% of the grasps were successful and 98\% of the placements were successful, resulting in a complete task pick/place success rate of 97\%. Column two shows similar results for the bottles-in-clutter scenario, and columns three and four include results for the same experiments with mugs. Finally, column five summarizes results from the multi-step-isolation scenario for mugs: overall, our method succeeded in placing the mug upright into the box 68\% of the time. The temporary place success is perfect because a temporary placement only fails if the mug is so high it rolls away after dropped or too low it is pushed into the table, neither of which ever happened after 72 trials. The final placement is perfect because it always did get the orientation right (for all 72 trials that got far enough to reach the final placement), and it is hard for the mug to fall over in the box. The multi-step scenario has low task success rate because 12 trials failed to perform the final place after $10$ time steps. Perhaps this is due to lower Q-function values on the real system (due to domain transfer issues), causing the robot to never become confident enough with its given state information to perform the final place.

Our experimental results are interesting for several reasons beyond demonstrating that the method can work. First, we noticed consistently lower place performance for the mug category relative to the bottle category. The reason for this is there is more perceptual ambiguity involved in determining the orientation of a mug compared to that of a bottle. In order to decide which end of a mug is ``up'', it is necessary for the sensor to view into at least one end of the mug. Second, the robot had trouble completing the multi-step task in a reasonable number of steps with the real hardware compared with simulation. This may be because fewer grasps are available on the real robot versus the simulated robot due to collision modelling. Another unexpected result was our learned policies typically prefer particular types of grasps, e.g., to grasp bottles near the bottom (see Figure~\ref{fig:robotExperimentIllustration}). We suspect this is a result of the link between the location of a selected grasp and the grasp descriptor used to represent state. In order to increase the likelihood that the agent will make high-reward decisions in the future, it selects a grasp descriptor that enables it to easily determine the pose of the object. In the case of bottles, descriptors near the base of the bottle best enable it to determine which end is ``up''.


\section{Conclusion and Future Work}

This paper proposes a new way of structuring robotic pick-place and regrasping tasks as a deep RL problem. Importantly, our learned policies can place objects very accurately without using shape primitives or attempting to model object geometry in any way. Our key insight is to encode a sampled set of end-to-end reaching actions using descriptors that encode the geometry of the reach target pose. We encode state as a history of recent actions and observations. The resulting policies, which are learned in simulation, simultaneously perceive relevant features in the environment and plan the appropriate grasp and place actions in order to achieve task goals. Our experiments show that the method consistently outperforms a baseline method based on shape primitives.

For future work, we plan to generalize the descriptor-based MDP in two ways. First, place poses could be sampled from a continuous, 6-DOF space, as grasps are. To do this we would develop a special purpose place detector in the same way GPD is a grasp detector. Second, the system should be able to work with a more diverse set of objects, e.g., kitchen items. This may require a CNN with more capacity and longer training time, motivating innovations to speed up the learning in simulation.


\bibliographystyle{IEEEtran}
\bibliography{References}

\end{document}